\documentclass[10pt,twocolumn,letterpaper]{article}

\usepackage{cvpr}
\usepackage{times}
\usepackage{epsfig}
\usepackage{graphicx}
\usepackage{amsmath}
\usepackage{amssymb}

\usepackage{float}
\usepackage{balance}
\usepackage{tablefootnote}
\usepackage[flushleft]{threeparttable}
\usepackage{multirow}
\usepackage[linesnumbered,ruled,vlined]{algorithm2e}
\usepackage{url}            
\usepackage{booktabs}       
\usepackage{amsfonts}       
\SetKwInput{kwSource}{Source}
\SetKwInput{kwProced}{Procedure}
\usepackage{amsmath,amssymb}
\usepackage{balance}

\DeclareMathOperator{\var}{\mathrm{Var}}

\usepackage[breaklinks=true,letterpaper=true,colorlinks,bookmarks=false]{hyperref}

\cvprfinalcopy 

\def\cvprPaperID{7890} 

\ifcvprfinal\fi
\begin{document}

\title{Regularizing CNN Transfer Learning with Randomised Regression}

\author{Yang Zhong \qquad Atsuto Maki\\
Division of Robotics, Perception, and Learning \\  
KTH Royal Institute of Technology, Sweden\\
{\tt\small \{yzhong, atsuto\}@kth.se}
}

\maketitle
\vspace*{-.9cm}
\begin{abstract}
This paper is about regularizing deep convolutional networks (CNNs) based on an adaptive framework for transfer learning with limited training data in the target domain.
Recent advances of CNN regularization in this context are commonly due to the use of additional regularization objectives.
They guide the training away from the target task using some forms of concrete tasks. 
Unlike those related approaches, we suggest that an objective without a concrete goal can still serve well as a regularizer. 
In particular, we demonstrate {\it Pseudo-task Regularization} (PtR) which dynamically regularizes a network by simply attempting to regress image representations to pseudo-regression targets during fine-tuning. 
That is, a CNN is efficiently regularized without additional resources of data or prior domain expertise. 
In sum, the proposed PtR provides:  
a) an alternative for network regularization without dependence on the design of concrete regularization objectives or extra annotations; 
b) a dynamically adjusted and maintained strength of regularization effect by balancing the gradient norms between objectives on-line.
Through numerous experiments, surprisingly, 
the improvements on classification accuracy by PtR are shown greater or on a par to the recent state-of-the-art methods.
\end{abstract}

\section{Introduction}
\label{sec:introdu}

Deep convolutional neural networks (CNNs) have recently advanced the development of computer vision and flourished in many large-scale computer vision applications \cite{imagenet_cvpr09,MScoco,CelebA,posenet}. 
Since the introduction of AlexNet \cite{AlexNet}, deeper and more complex network architectures, such as VGG \cite{VGG}, Inception \cite{goingdeeper}, 
ResNet \cite{Resnet}, and DenseNet \cite{densenet},
have been proposed. 
In addition, other contributions have been made toward network optimization, which has been helping the performance and efficiency of CNNs, e.g. 
BatchNorm \cite{batchnorm} and MiniBatchSGD \cite{miniSGD}.   
Despite the improved effectiveness by those, one of the known open issues is that CNNs are normally over-parameterized and would demand a large-scale labeled dataset.

It is a common practice to exploit transfer learning which adapts a model pre-trained on a source task to a new target task when given a small amount of labeled dataset.  
Specifically, by leveraging the transferability of deep features \cite{NIPS2014_5347} one can 
map images to a middle or high-level feature through pre-trained model and therewith train target specific classifiers \cite{pmlr-v32-donahue14,Oquab14,Zeiler2014}, which is often called {\it feature selection}.  
It is also viable to {\it fine-tune} a source model for the target data. 
As fine-tuning aims to optimize the entire network for the target task, it often achieves higher effectiveness and has therefore been a rule of thumb 
in CNN transfer learning with a limited amount of domain data \cite{7328311}.
During fine-tuning, a source model needs to be mildly tuned to avoid overfitting due to the fact that deep networks are yet over-parameterized for small-scale target tasks. 

One of the challenges during fine-tuning, which this paper also addresses, is to achieve network regularization for an over-parameterized model with limited training samples.
In the recent state-of-the-art transfer learning solutions, there is a trend of using an auxiliary training objective in a framework of multi-objective\footnote{For ease of discussions, we do not distinguish training ``objectives'' from ``tasks''; thus, multi-objective and multi-task learning may be used interchangeably.} learning for improved regularization \cite{Borrow,LwF,IndB,PC}.
These auxiliary objectives are designed in a concrete and target-specific manner, through which models would enforce certain desired properties that facilitates multiple purposes in the learned image representations.
The key to the enhanced regularization on the target task is then attributed to the improved generality learned from the imposed auxiliary objectives through partial or entire source task data. 
However, the regularization gain comes with a resource-dependent cost of the storage of off-the-shelf predictions for multiple steps of network training \cite{LwF}, selecting qualified labeled data samples from the source domain for a target task \cite{Borrow}, using a complex network architecture during training \cite{PC}, or recalling the source model \cite{IndB}.

From a network training perspective, as another way to understand the regularization, a basic effect of training with a regularization objective could be considered as to 
\textit{distract} the minimization of the empirical loss (typically, through a structural loss).  
As a result, the regularization power can also be seen to come from the extra gradients generated by the employed distracting (regularization) objective.
These gradients cause useful distortions in the gradient-descent trajectory to force the network to tolerate slightly higher empirical loss in the course of training, which allows for more chances in seeking better optima. 
Now, if such a distraction effect is the essence to network regularization, it is worthwhile to study whether the regularization objective could have some alternative form rather than being a real and concrete task. 

Intuitively, if it is the distraction (rather than the convergence of a regularization task) that is the primary interest, there could be diverse ways to construct a \textit{distractor} which interferes the training of the target task while looking for an improved regularization.
One potential approach could be through a {\it pseudo-task} which neither depends on the above mentioned data and storage availability for multi-task learning, nor a concrete goal as designed in \cite{Borrow,LwF,IndB,PC}. 

In this paper, 
considering image classification tasks in a transfer learning scenario,
we aim to device a regularizer which generates distractions while being independent of concrete tasks.
Our regularizer simply exploits a {\it pseudo-task}\footnote{Networks are never able to converge on the pseudo-task as it leverages random regression targets, hence the name pseudo-task.
See the details in Sec. \ref{sec:algo}.} 
that injects random noise in the gradients to distract the training on the target tasks, 
to seek for improvements.
Experimental results consistently support our conjecture on various datasets and with different network architectures. 
The contributions of this paper are:
\begin{enumerate}
\item We demonstrate \textit{Pseudo-task Regularization} (\textbf{PtR}) that provides an efficient alternative to other recent best regularization based on real and concrete tasks. 

\item In PtR, useful gradients for regularization are generated through a pseudo-task, in which we propose to dynamically adjust the strength of the regularization based on the gradient norms of the target objective and the pseudo-task. 
\end{enumerate}
The results by those suggest a novel interpretation
on the key elements of network regularization for CNN transfer learning, which we hope future research will exploit further to identify the essential requirements for a regularizer.

\begin{figure}[tb!]
\centering
\includegraphics[width=0.47\textwidth]{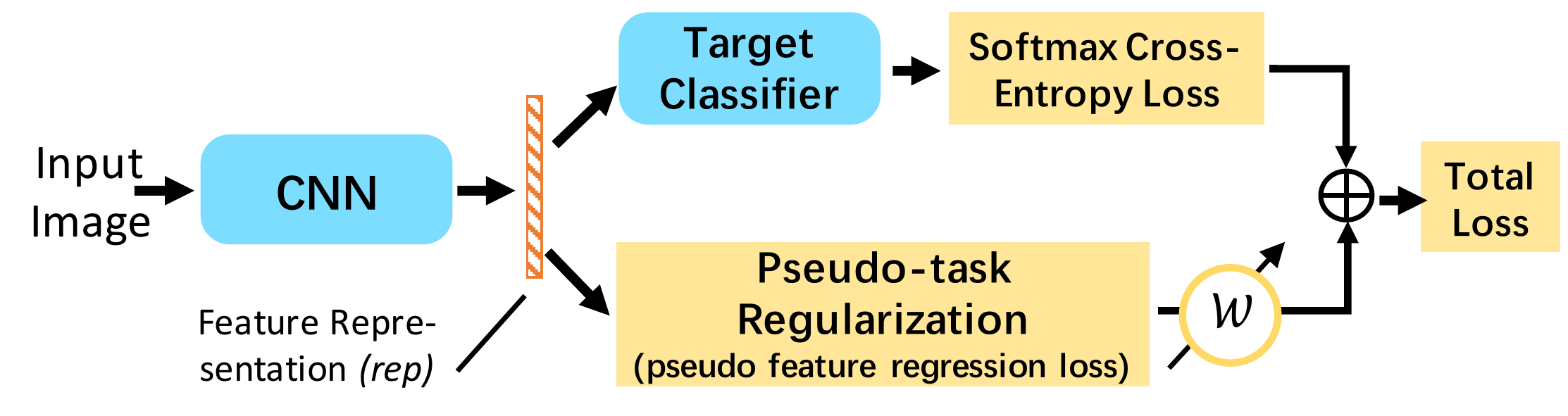}
\caption{An overview of the proposed Pseudo-task Regularization (PtR). 
The path linking the blue modules illustrates a vanilla fine-tuning pipeline with a target classifier trained by a cross-entropy loss.
The PtR loss, which is connected to the feature representation layer of a CNN, is brought into the network training when the convergence on target task is relatively stable.
The total loss therewith is the sum of cross-entropy loss on the target task and the weighted PtR loss.
The PtR loss module automatically weights the strength of regularization according to the gradient norm of the target task on the feature layer. 
The PtR is explained in Algorithm \ref{algo:1} in detail.}
\label{fig:network}
\end{figure}

\section{Pseudo-task Regularization}
\label{sec:approach}
\subsection{Overview}

Our motivation is to let a CNN learn the representations for a target task while also being distracted so that the learned representations are not excessively target-task specific, leading to the loss of generality.
For this purpose, as shown in Figure \ref{fig:network}, we choose to exploit a multi-task learning framework which utilizes two training objectives: 
one is the cross-entropy loss for the target-task classification, 
the other objective generates distractions to promote the generalization through a pseudo-regression task. 
We call it Pseudo-task Regularization (PtR).
 
In using both of the training objectives, a significant aspect of PtR is to balance the impact of the two loss functions. 
It is reasonable that the distraction should be on a proper level which is not too strong and hinder model convergence, nor is on an ignorable magnitude compared to the gradients of the target classifier. 
To this end, inspired by \cite{GNorm}, we propose to dynamically balance the strengths of gradients from the two losses according to the gradient norms with respect to the image representation during the training process. 
The training procedure of our adaptive multi-task learning framework is described in Algorithm \ref{algo:1}.

\subsection{Algorithm}
\label{sec:algo}

\begin{algorithm} 
\SetAlgoLined
\caption{Training with Pseudo-task Regularization}
\label{algo:1}
 \kwSource{ a) Off-the-shelf net; b) Labeled data in target domain}
 \kwProced{\\
  \For{iteration (batch) $i$ } {
  Compute cross-entropy loss $L^{(i)}_{ce}$.  \\
  	\eIf{ $L^{(i)}_{ce} $ far from minimum} 
    	{Back propagate $L^{(i)}_{ce}$ only;}
    	{
        First, perform the following calculations:

        1. $L^{(i)}_{PtR}$ : the pseudo-regression task loss, w.r.t. the regression target $\it{t}^{(i)}$ generated on-line;
 
        2. $G^{(i)}_{ce}$ and $G^{(i)}_{PtR}$ : the gradient norms of  $L^{(i)}_{ce}$ and $L^{(i)}_{PtR}$ w.r.t. $rep^{(i)}$. $rep^{(i)}$ stands for the image representations of the batch;

        3. $\overline{G}^{(i)}_{ce}$ and $\overline{G}^{(i)}_{PtR}$ : the average of $G^{(i)}_{ce}$ and $G^{(i)}_{PtR}$ over the batch; 

        4. Weight $w$: $w$ =  $\frac{\overline{G}^{(i)}_{ce}}{\overline{G}^{(i)}_{PtR} \cdot R}$ for a target ratio $R$. 
        
        Then, back propagate $L^{(i)}_{ce} + w \cdot L^{(i)}_{PtR}$.
        
         }
    	
  }
  }

\end{algorithm}

Our method learns image representations on a target task using a pre-trained model in an adaptive multi-objective learning framework, as shown in Algorithm \ref{algo:1}. 
For a training iteration $i$, it computes the cross-entropy loss $L^{(i)}_{ce}$ and additionally the loss by the randomised regressor, the {\it Pseudo-task Regularization loss} $L^{(i)}_{PtR}$, starting when $L^{(i)}_{ce}$ falls below a certain threshold of average epoch loss $T$.
(The choice of the threshold is not critical as explained in Figure 2 of the supplementary.)
$L^{(i)}_{PtR}$ is calculated by regressing the image representation to a pseudo-regression target.
In PtR, we use random regression targets generated on-line such that:
\begin{equation}	
L^{(i)}_{PtR} = f_{reg}(\it{rep}^{(i)}, \it{t}^{(i)}),
\end{equation}
where $\it{rep}$ stands for the activation of the representation layer in a CNN, $t$ for pseudo-regression targets with an equal dimension as $\it{rep}$, and $f_{reg}(\cdot)$ is a regression function 
for which we consider two popular choices: $L2$ loss and ``smooth-L1'' (denoted by SML1) loss. 
Note that the pseudo targets are randomly generated during training so that the training instances are not bounded to the generated regression targets. 
The details of these targets are described in Section \ref{sec:pseudo-regression-tasks}.

The total loss $L^{(i)}_{total}$ 
is the weighted sum of the cross-entropy loss and the regression loss:
\begin{equation}
L^{(i)}_{total} = L^{(i)}_{ce} + w L^{(i)}_{PtR}, 
\vspace{-.2em}
\end{equation}
where $w$ is a coefficient to balance the impact of the distraction regressor, as explained below; $L^{(i)}_{ce}$ and the balanced regression loss $w \cdot L^{(i)}_{PtR}$ are back propagated through the network.
Weight decay is omitted for conciseness.

To generate a proper level of distraction for regularization, we first calculate the gradient norms of the cross-entropy loss and those of the regression loss w.r.t. the output feature for each instance, which are denoted by $G^{(i)}_{ce}$ and $G^{(i)}_{PtR}$ , respectively (for brevity, the indexes of the training instance in batch $i$ are omitted): 
\begin{equation}
G^{(i)}_{ce} =||\frac{\partial{L^{(i)}_{ce}}}{\partial{rep^{(i)}}}||_{2},\hspace{3mm}  
G^{(i)}_{PtR} = ||\frac{\partial{L^{(i)}_{PtR}}}{\partial{rep^{(i)}}}||_{2}.
\end{equation}
The gradient norms are then averaged over the batch as:
\begin{equation}
\overline{G}^{(i)}_{ce} = \mathbb{E}[G^{(i)}_{ce}], \hspace{3mm}  
\overline{G}^{(i)}_{PtR} = \mathbb{E}[G^{(i)}_{PtR}].
\end{equation}
In order to balance the relative impact of $L^{(i)}_{ce}$ and $L^{(i)}_{PtR}$ , we introduce a target gradient norm ratio $R$.
It is defined by the gradient norm ratio of the cross-entropy loss and a desired regression loss in the form of signal-to-noise ratio: $R =  {\overline{G}^{(i)}_{ce}} / w {\overline{G}^{(i)}_{PtR}}$.
Thus, for the gradient norm ratio to satisfy $R$ at iteration $i$, $L^{(i)}_{PtR}$ needs to be weighted by a factor $w$:
\begin{equation}
w = \frac{\overline{G}^{(i)}_{ce}}{\overline{G}^{(i)}_{PtR} \cdot R},
\end{equation}
which is calculated on-line per batch before back propagation. 

It is worth nothing that the common way of balancing losses through a fixed weight ($w$) would not be an efficient design choice for PtR, as it would never guarantee precise regulation on the gradient norm for regularization. 
On the contrary, with the use of a dynamic weight PtR is not bound to a particular regularization (regression) loss. 


\subsection{Encouraging higher variance in gradients}  
\label{sec:vargrad}
To explore the impact to the gradients by using a randomly varying regression target, we case-study a minimum toy example network which is composed by one hidden neuron (with non-linear activation) connected to one input and one output.  
The single hidden neuron is denoted by $f$, whose output $f^{(o)}$ is seen as the feature representation learned by the example network.
The input to $f$, denoted by $f^{(i)}$, is the product of an input $x$ and its learnable weight $a$ on the input path. 
That is: $f^{(o)} = \delta(f^{(i)})$, $f^{(i)}$ = $a \cdot x$, and $\delta(\cdot)$ represents the ReLU function.

When a regression target $t$ is applied to $f^{(o)}$, the regression loss 
$E_{reg} = \frac{1}{2} (f^{(o)} - t)^{2}. $
During back propagation, if neuron $f$ is activated, the gradient on $a$ by the chain rule is:
\begin{equation}
\label{eq:v2}
\begin{split}
    \frac{\partial E_{reg}}{\partial a} & = \frac{\partial E_{reg}}{\partial f^{(o)}}   
                             \cdot             \frac{\partial f^{(o)}}{\partial f^{(i)}} 
                              \cdot            \frac{\partial  f^{(i)}}{\partial a}
                                        = \mid f^{(o)} - t \mid \cdot \it{x} \\
                                        & =
                                        \begin{cases}
                                        (f^{(o)} -t) \cdot \it{x}, &\text{if } (f^{(o)} -t) \geq 0, \\
                                        -(f^{(o)} -t) \cdot \it{x}, &\text{otherwise}.
                                        \end{cases}
\end{split}
\end{equation}
The variance of the gradient of $a$, if a simplified assumption can hold that $x$ being a constant, $Var(\frac{\partial E_{reg}}{\partial a})$, is determined by that of $f^{(o)}$ and regression target $t$, such that 
\begin{equation}
\label{eq:v3}
    \var(\frac{\partial E_{reg}}{\partial a}) = \var(f^{(o)}) + \var(t), 
\end{equation}
given that $t$ is a variable independent of $f^{(o)}$.

It can be seen that, if the regularization is achieved through feature norm penalization (e.g., as proposed by \cite{fewshot}), $t$ in Equation \ref{eq:v2} equals a constant of 0. 
Consequently, in Equation \ref{eq:v3}, the variance of gradients gets smaller compared to using other regression targets which follow a certain distribution.  
PtR generates gradients with higher variance using an independent randomly-varying pseudo target. 
By leveraging them, PtR would explore more local optima to yield higher chances in avoiding saddle points and achieving stronger regularization.

\section{Experiments and Results}

\subsection{Experimental Setup}
\label{sec:pseudo-regression-tasks}

\textbf{Datasets.}
For transfer learning, training CNNs with data across domains has been found to be an important regularization method. 
However, our experiments focus on a situation where data of other domain is not available.
We also focus on a challenging scenario where the training samples are  sparse.
For this purpose, four commonly used small-scale transfer learning datasets are selected to comparatively evaluate PtR: Flower102 \cite{Flowers}, CUB200-2011 \cite{WahCUB_200_2011}, MIT67 \cite{MITindoor}, and Stanford40 \cite{Stanford40}, two of which represent fine-grained classification tasks of different scenarios.
Besides, we also chose 500 identities\footnote{Random 500 identities that have the most training instances on the WebFace dataset.} from the WebFace \cite{webface} dataset, denoted by ``WebFace500'', to evaluate PtR when performing transfer learning from image classification to closed-set face identification with scarce samples per class.  
Caltech256 \cite{caltek} was also used for performance evaluation in a general image recognition scenario.

On Flower102 we faithfully follow the data splits for training and testing. 
On the WebFace500 dataset, each identity has random 20 training images, five validation images, and on average 24 test images. 
Faces were segmented and normalized to a fixed scale with a face detector \cite{facedet} before training.
From Caltech256, we formed two independent training sets with 30 and 60 training samples per class, respectively, for consistency with \cite{Borrow,IndB}. 
For other datasets, $10\%$ of the training images were randomly separated to form the validation sets for model training. 

\textbf{Training and Evaluation.}
To augment the training images, we employed random jittering and subtle scaling and rotation perturbations to the training images.
We resized images of all involved datasets to $250 \times 250$ pixels, and the aspect ratio of the images was retained by applying zero padding all the time.
During test time, we averaged over the network responses from the target-task classifiers over ten crops which were sampled from the corners and the centers of originals and the flipped counterparts.

As we consider the vanilla fine-tuning procedure as the baseline, it is very important to ensure that the effectiveness of vanilla fine-tuning is not underestimated. 
To this end, we carefully selected learning rate schedules for fine-tuning to demonstrate the test accuracy on each dataset with each type of network architecture. 
To conduct fair comparisons to fine-tuning as much as possible, we also used the same learning rates used by fine-tuning in our dynamic pseudo-task regularization approach; 
the learning rate schedules were slightly different due to the difference in converging speeds. 
A learning rate was decreased when the validation loss and validation accuracy stopped progressing and it was decreased twice before model training was terminated. 
The models trained after the last epoch of their learning rate schedules were always used for performance evaluations. 

\textbf{Implementation details.} Experiments on different datasets shared many common settings. 
We used the standard SGD optimizer with momentum set to 0.9.
The batch size was set to 20 to reduce overfitting as much as possible (unless otherwise stated); weight decay was set to 0.0005 for VGG networks \cite{VGG} and 0.0001 for ResNet \cite{Resnet} architectures except in a number of ablation studies. 
The dropout ratio for VGG networks was set to 0.5. 
Our experiments were implemented with PyTorch \cite{pytorch}.

We always started the experiments from an ImageNet \cite{imagenet_cvpr09} pretrained model.
As the training data was visited randomly, we ran \textbf{five independent runs} and average the results to mitigate the impact of randomness for all the experiments.
The classification accuracy was mostly used to compare with related methods except \cite{Borrow}.

\textbf{Other hyperparameters to PtR.} In the PtR, the impact of the additional loss is adjusted primarily by the target gradient norm ratio $R$ that controls the interference gradient magnitude. 
Then, the gradients with respect to each feature dimension are largely determined by the nature of a pseudo-task which we employ in our experiments, such as the distribution of pseudo-regression targets. 
Without loss of generality we considered the random targets, $\it{t}^{(i)}$, following a uniform distribution with a mean value of $m$ such that
$m = \mathbb{E}[\it{t}^{(i)}_{j}];$ 
i.e., for any single regression target $\it{t}^{(i)}_{j}$ in $\it{t}^{(i)}$,
$\it{t}^{(i)}_{j} \in [0, \text{2}m)$, where $j$ is an instance in batch $i$ ($j \in [1,|i|]$) and $|i|$ is the batch size.

We used independent hold-out sets to efficiently determine $R$ and $m$ (to avoid expensive cross-validation parameter search):  
for ResNet structures $R$=3 and $m$=1 are consistently used on all datasets;
for the VGG-16 structure, $R$ varies in the range between 3 and 5, and $m$ around 10 to 15. 
We chose $T$=1 as a sensible setting in all the experiments given that the influence by the choice of $T$ is limited.
Details are given in Figure 2 of the supplementary.

\begin{table*}[!bt]
\centering
\small
\caption{The comparative classification accuracy by the Pseudo-task Regularization (PtR), against vanilla fine-tuning (in column Baseline) with two different choices of regression functions, SML1 and L2, with the VGG-16 architecture. The performance gain brought about by PtR with different regression loss functions are in the two columns in the center under SML1 and L2, respectively. The corresponding error rate reduction values are in the rightmost columns. The standard deviation of each experiment is given in parenthesis.}
\vspace{.3em}
\label{res:v16}
\begin{footnotesize}
\begin{tabular}{cccccc}
\toprule
\multirow{2}{*}{} & \multirow{2}{*}{Baseline} & \multicolumn{2}{c}{Regularization Gain} & \multicolumn{2}{c}{Error Rate Reduction} \\ \cline{3-6}
                  &                  & SML1          		& L2           		& SML1        & L2                                    \\ \midrule
Flower102         & 83.92\% (0.36)   & 2.38\% (0.32)       & 2.61\% (0.42)        & 14.80\%     & 16.23\%                                      \\ 
CUB200            & 75.07\% (0.26)   & 3.05\% (0.39)       & 2.84\% (0.37)        & 12.23\%     & 11.39\%                                      \\ 
MIT67             & 71.55\% (0.38)   & 1.42\% (0.58)       & 1.39\% (0.40)        & 4.99\%      & 4.89\%                                      \\ 
Stanford40        & 76.99\% (0.19)   & 2.50\% (0.09)       & 2.21\% (0.16)        & 10.86\%     & 9.60\%                                      \\ 
WebFace500         & 77.54\% (0.52)   & 0.95\% (0.56)       & 0.83\% (0.47)        & 4.23\%      & 3.70\%                                      \\  \bottomrule
\end{tabular}
\end{footnotesize}
\vspace{-0.5em}
\end{table*}

\subsection{Results and Comparisons}

As the VGG-16 architecture has been used very commonly in many different transfer learning applications, we first evaluate PtR with it across five different dataset using SML1 and L2 for the regression function respectively, and compare against the fine-tuning baseline. 
The results are listed in Table \ref{res:v16}.

It is observed that PtR helps improve the vanilla fine-tuning on all the classification tasks tested with two different regression functions; it brings about reasonable and consistent performance gain.
On WebFace500 which contains the most training samples, it reduced the error rate for around $4\%$, but on Flower102 and CUB200 where the training samples are more sparse, it is particularly more effective and reduces the error rates for more than 10\%. 
These results suggest that, on the one hand, 
collecting more data helps regularization even for small datasets.
On the other hand, when the training samples become sparse, PtR manages to keep the learned representations from being excessively target-task specific and further promises the networks to learn more useful representations.
The choice of regression function does not appear to be a significant factor as the test accuracy with SML1 and L2 are close; SML1 is used in all the following experiments.

We have also conducted a large number of comparative experiments to recent best performing multi-task/objective based regularization approaches: 
Joint Training (\textbf{JointTrain}) \cite{LwF}, 
Learning without Forgetting (\textbf{LwF}) \cite{LwF}, 
Borrowing Treasures from the Wealthy (\textbf{BTfW}) \cite{Borrow}, 
Inductive Bias (\textbf{Ind.Bias}) \cite{IndB}, and Pair-wise Confusion (\textbf{PC}) \cite{PC}.
In addition, we evaluated the regularization through feature norm penalty (denoted by \textbf{FNP}) \cite{fewshot} in the context of transfer learning (hyperparameters of FNP were therefore set by using the same procedure of PtR for fairness).
We have also compared the impact of weight decay on CUB200 and Caltech256 by disabling weight decay (denoted by ``w/o WD'').
As we intended to perform all experiments on a single GPU module with 12GB memory, the ResNet-101 was used as a compromise to compare to the results achieved by a special memory-saving version of ResNet-152 in \cite{Borrow}.
For fair evaluations, all the corresponding vanilla fine-tuning baseline and improved test accuracy are shown in the following tables together with accuracy gain.

The comparative results on CUB200 dataset are shown in Table \ref{tab:cmpcub}.  
It can be seen that the accuracy gain by PtR (with VGG-16) is slightly better than JointTrain where real source data was used for regularization;
it also performs better than LwF where off-the-shelf predictions were used. 
Compared to FNP, PtR outperforms by a visible margin of 1.3\%. 
Although PtR achieved lower accuracy gain than PC, the gaps is not significant regardless of network architecture (around 0.5\%).
For the absolute accuracy, it is also noticeable that PtR achieved the highest baseline performance as well as that of the optimized models among all the other methods.
Weight decay seems not impacting PtR, but the baseline accuracy is 0.7\% higher when training without it.

\begin{table}[htbp]
\centering
\small
\caption{Comparing classification accuracy on the CUB200 dataset. All the numerical results are in \%. The network and other settings (if any) used by each method are given in parenthesis.}
\vspace{.3em}
\begin{footnotesize}
\begin{tabular}{cccc}
\toprule
Method               & Baseline &  Acc. 	& Gain  \\ \midrule
\multicolumn{1}{r}{JointTrain (VGG-16)} & 72.1     & 74.6		& 2.5	\\ 
\multicolumn{1}{r}{LwF (VGG-16)}        & 72.1     & 72.3		& 0.2   \\ 
\multicolumn{1}{r}{PC(VGG-16)}          &  73.3    & 76.5		& \textbf{3.2 }   \\ \hline 
\multicolumn{1}{r}{{\bf PtR} (VGG-16)}  &  75.1    & 78.1		& 3.0      \\ \midrule 
\multicolumn{1}{l}{PC (ResNet-50)}      &  78.2    & 80.3		& \textbf{2.1}    \\
\multicolumn{1}{l}{FNP (ResNet-50)}     &  80.3    & 80.6  	    & 0.3             \\ \hline
\multicolumn{1}{l}{{\bf PtR} (ResNet-50)} &  80.3  & 81.9			& 1.6     \\
\multicolumn{1}{l}{{\bf PtR} (ResNet-50, w/o WD)} &  81.0  & 82.0			& 1.0     \\ \bottomrule
\end{tabular}
\label{tab:cmpcub}
\end{footnotesize}
\end{table}

On Flower102 dataset, as shown in Table \ref{tab:cmpflow}, the gain by PtR is larger than PC for 1.4\% with the VGG-16 structure; it is just equivalent to that of PC with the ResNet-50 network. 
FNP brings some regularization margin, but it is 0.3\% lower than that of PtR.
For PtR, it achieves consistent gain in accuracy with ResNet-50 and ResNet-101, and the depth of the network does not appear to deteriorate the regularization effect. 
Although we achieved an equally good baseline performance as BTfW (in parenthesis of the bottom row of Table \ref{tab:cmpflow}), the regularization gain of BTfW is higher than PtR or any other methods. 
The difference in regularization may suggest that training with sufficient labeled data in a multi-task learning framework is a stronger regularization for transfer learning.

\begin{table}[htbp]
\centering
\small
\caption{Classification accuracy and accuracy gain (in \%) comparisons on the Flower102 dataset. The mean class accuracies used to compare to BTfW \cite{Borrow} are listed in parenthesis.}
\vspace{.3em}
\begin{footnotesize}
\begin{tabular}{cccc}
\toprule
Method             							& Baseline 			& Acc. 					& Gain \\ \midrule
\multicolumn{1}{r}{PC(VGG-16)}         		& 85.2    			&	86.2				& 1.0      \\ \hline
\multicolumn{1}{r}{{\bf PtR} (VGG-16)}       	& 83.9    			&	86.3		& \textbf{2.4}      \\ \midrule
\multicolumn{1}{l}{BTfW (ResNet-152)} 	& 92.3     			&	94.7		& \textbf{2.4}	\\ 
\multicolumn{1}{l}{PC (ResNet-50)}     		& 92.5	     &	93.5		& 1.0      \\ 
\multicolumn{1}{l}{FNP (ResNet-50)}        &  91.0      & 91.5  		& 0.5      \\     \hline
\multicolumn{1}{l}{{\bf PtR} (ResNet-50)}       	& 91.0     			&	91.8				& 0.8      \\ 
\multicolumn{1}{l}{{\bf PtR} (ResNet-101)}       & 90.6(92.3)    	&	91.6(93.2)			& 1.0(0.9)      \\ \bottomrule
\end{tabular}
\label{tab:cmpflow}
\end{footnotesize}
\end{table}

\begin{table}[thbp]
\centering
\small
\caption{Comparative results on the MIT67 (in \%). The mean class accuracies used to compare to BTfW \cite{Borrow} are in parenthesis.}
\vspace{.3em}
\begin{footnotesize}
\begin{tabular}{cccc}
\toprule
Method                & Baseline & Acc. 	& Gain \\ \midrule
\multicolumn{1}{r}{JointTrain (VGG-16)}     & 74       &	75.5	& \textbf{1.5}  \\ 
\multicolumn{1}{r}{LwF (VGG-16)}            & 74       &	74.7	& 0.7        \\ \hline
\multicolumn{1}{r}{{\bf PtR} (VGG-16)}         	& 71.6    	&	73.0	& 1.4      \\ \midrule
\multicolumn{1}{l}{BTfW (ResNet-152)}     & 81.7     &	82.8	& \textbf{1.1} 		\\ 
\multicolumn{1}{l}{Ind.Bias (ResNet-101)}   & 77.5     &	78		& 0.5      \\ 
\multicolumn{1}{l}{FNP (ResNet-50)}        & 77.4     & 78.0  		& 0.6      \\    \hline
\multicolumn{1}{l}{{\bf PtR} (ResNet-50)}      	& 77.4     	& 77.9		& 0.5	    \\ 
\multicolumn{1}{l}{{\bf PtR} (ResNet-101)}      & 78.7(78.7)    & 79.2(79.2)	& 0.5(0.5)     \\ \bottomrule
\end{tabular}
\label{tab:mit}
\vspace{-.5em}
\end{footnotesize}
\end{table}

Similar observations can also be found from the results on MIT67 dataset, as shown in Table \ref{tab:mit}.
With the VGG-16 architecture, regularization effect by PtR is again very close to that of JointTrain and outperforms LwF. 
With the ResNet, the regularization gain by PtR is equivalent to that of Ind.Bias and FNP.
BTfW also achieves higher gain than the other methods with ResNet.
We would infer that optimizing a network simultaneously on multiple tasks with sufficient selected real data samples might be more effective than other related methods. 
As for PtR, it brings about consistent margin over the fine-tuning base-line regardless of the depth of ResNet architecture, which also coincides with the results in Table \ref{tab:cmpflow}.

The results on Caltech256 dataset are in Table \ref{tab:cmpcaltk}.
In these experiments, we increased the batch size to 32, which is a value in between of those used by \cite{Borrow,IndB}, to make fair comparisons as much as possible.
Interestingly, we achieved the best baseline accuracy among all the comparing methods with both of the Caltech256 partitions.
Consequently, it could be harder for PtR to demonstrate the regularization power in comparison to others because a better generalized baseline usually has a smaller room to improve the generalization. 
However, we can still observe some similar trends.
First, as in the previous experiments, by training a network with sufficient annotated data of multiple classes, BTfW achieves the best regularization gain (around 2.6\% with both setups).
Second, PtR consistently delivers regularization gain; for both data splits the gains are equivalent, which indicates that PtR would not be so sensitive to the size of training data of each category. 
The improvement brought about by FNP might be marginal or even unstable given the negative gain on Caltech256-30.
The impact of weight decay on the classification accuracy of PtR is not visible.

\begin{table}[t!]
\centering
\small
\caption{Mean class accuracies and accuracy gains (in \%) on the Caltech256 with two partitions of training data. 
Bsln is short for baseline. 
Mean class accuracies were the same as the average classification accuracy over the test set, thus are not given in parenthesis in this table. 
BTfW \cite{Borrow} was using ResNet-152 while others were using ResNet-101. The network used by each method is not shown in this table for brevity.}
\vspace{.3em}
\begin{footnotesize}
\begin{tabular}{lccccccc}
\toprule
						& \multicolumn{3}{c}{Caltech256-30}  &  		& \multicolumn{3}{c}{Caltech256-60} \\ \midrule
Method                 & Bsln. 	& Acc. 	 		& Gain 				&  		& Bsln.       & Acc. 		& Gain \\ \hline
BTfW    & 81.2     	& 83.8     		& \textbf{2.6}  	&  		& 86.4           & 89.1    & \textbf{2.7}      \\ 
Ind.Bias & 81.5     	& 83.5     		& 2.0  				&  		& 85.3           & 86.4      		& 1.1     \\ 
\multicolumn{1}{l}{FNP}  &  84.0      & 83.8  		& -0.2    &    &  86.8      & 86.9  		& 0.1             \\  \midrule
{\bf PtR}               & 84.0 		&84.5			& 0.5   			& 		& 86.8			 & 87.2      		& 0.4     \\ 
{\bf PtR}\scriptsize{,w/o WD}        & 84.0 		&84.5			& 0.5   			& 		& 86.9			 & 87.2      		& 0.3     \\ \bottomrule
\end{tabular}
\label{tab:cmpcaltk}
\end{footnotesize}
\end{table}

Through the analysis of the comparative results, we argue that PtR delivers consistent gains, with statistical significance (also see the standard deviation given in Table 1 in the supplementary), which are on a par with the recent state-of-the-art approaches.
The compared methods consider using auxiliary objectives attached to concrete tasks while enhancing regularization, 
but PtR which leverages an additional pseudo-task as a regularizer is free from design of concrete auxiliary tasks and more straightforward. 
Compared to LwF, the ``warmup'' training stage and collecting predictions of the target data from an off-the-shelf model is not required by PtR. 
It is not needed for PtR to remember the ``starting point'' parameters of an off-the-shelf model as in \cite{IndB}.
PtR is also more efficient than \cite{PC} which requires a Siamese network, and it does not depend on annotated data from other domain either as in \cite{Borrow}.

\section{The Effect of PtR on Predictions}

\begin{figure*}[t]
\centering
\includegraphics[width=0.86\textwidth]{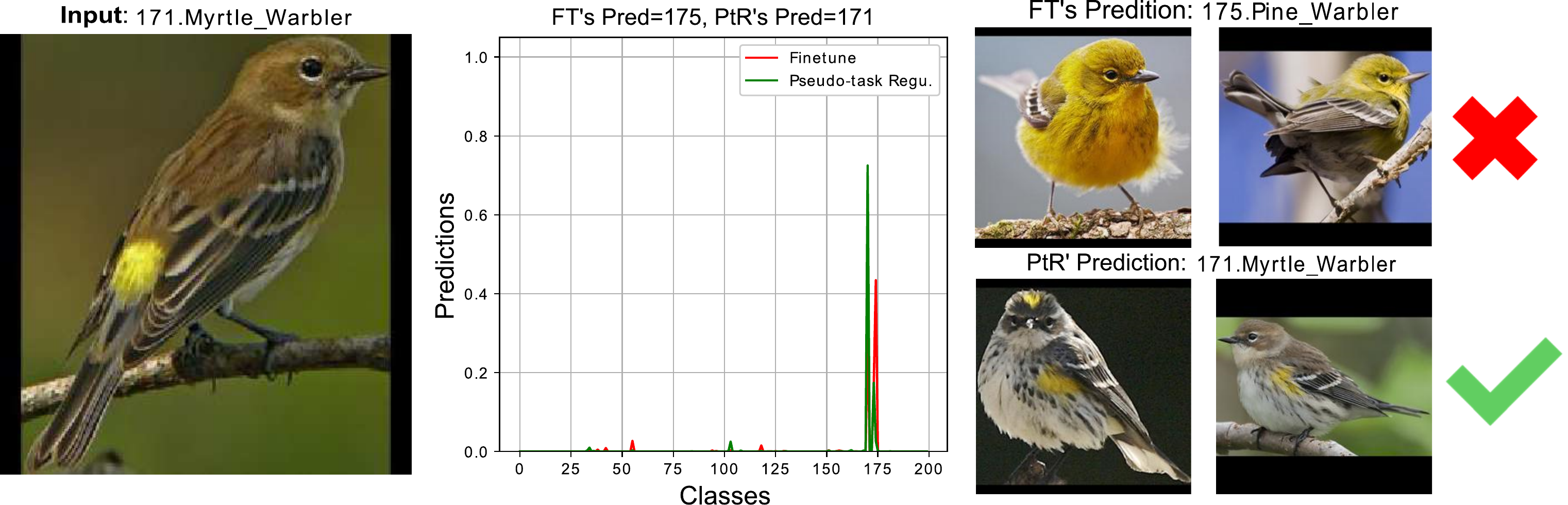}
\caption{ 
A sample from the validation set of CUB200 that PtR correctly rectified mis-classification caused in the vanilla fine-tuning.
Left: An input image. 
Middle: Categorical distributions made by the baseline model and PtR. The second largest prediction made by FT baseline model is around 30\% for Class 171.
Right: Two randomly picked training samples of the class predicted by the baseline (top) and two samples of the class predicted by PtR (bottom).
}
\label{fig:true}
\vspace{-1em}
\end{figure*}

To investigate the impact of PtR, we case-study on the validation set of the CUB200 dataset with ResNet-50 network to explore how the predictions have been altered in comparison to those from vanilla fine-tuning.
We base our analysis on the concept of confusion matrix and define a matrix ${\bf C}^{(D \times D)}$ (D=200), 
each row of which contains cumulative predicted probabilities across all the validation samples for different class categories.
We compute matrices ${\bf C}_{PtR}$ and ${\bf C}_{ft}$ for the cases of using PtR and baseline fine-tuned model, respectively.
We then sum their diagonal elements into:
\begin{equation}
S_{PtR} = \sum Diag({\bf C}_{PtR}), \hspace{3mm} S_{ft} = \sum Diag({\bf C}_{ft}).
\end{equation}
Feeding in 584 validation images, we had $S_{PtR}$ = 425 and $S_{ft}$ = 404, which indicates that the pseudo-task regularized model shows more certainty in the correct classes on average than the fine-tuned model.
Furthermore, in terms of the average entropy of the predictions, the pseudo-task regularizer reduced it from 1.33 to 1.15 bits. 
This is due to better regularization that allows the model to eliminate minor probabilities in false predictions, which in turn reduced the average entropy.
The reduction of entropy also implies that the class predictions have been disambiguated by PtR.

Correspondingly we computed $S'_{PtR}$ and $S'_{ft}$, the sum of off-diagonal elements of ${\bf C}_{PtR}$ and ${\bf C}_{ft}$, respectively. 
We observed that the regularized model tends to make predictions with fewer minor probabilities than the vanilla fine-tuning model.
It gives higher certainty to its predictions on the ground truth class given the smaller $S'_{PtR}$, which is consistent with the reduced entropy in comparison to vanilla fine-tuning.

\begin{figure*}[t]
\centering
\includegraphics[width=0.86\textwidth]{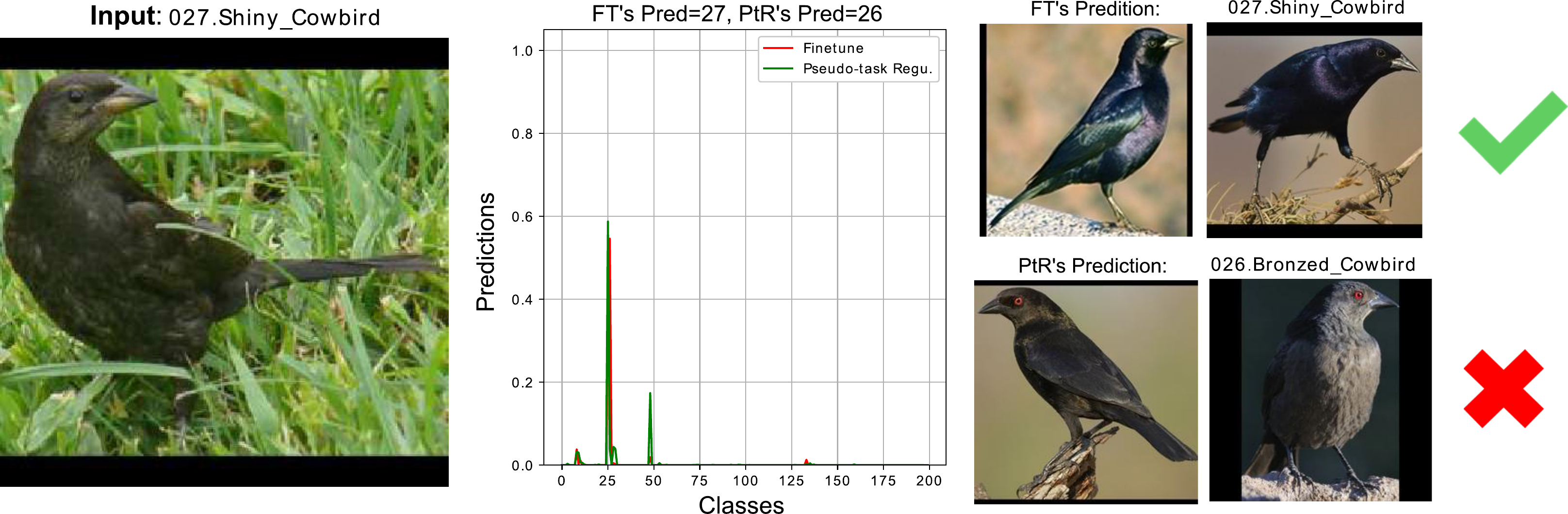}
\caption{
A sample from the validation set of CUB200 that the standard fine-tuning correctly classified but PtR wrongly predicted.
See also the caption of Figure \ref{fig:true}.}
\label{fig:false}
\vspace{-1.2em}
\end{figure*}

To further qualitatively study how the predictions made by the vanilla fine-tuning model (baseline) have evolved by PtR, we case study two types of input samples for which: \\
\hspace*{6mm} i) PtR rectified the baseline model's errors, and\\
\hspace*{5mm} ii) PtR mis-classified on the contrary to the baseline.\\ 
Namely, i) is {\it true rectification} and ii) is {\it false rectification}.
Two of these examples are compared in Figure \ref{fig:true} and \ref{fig:false}.

It can be seen that PtR has an effect of encouraging the predictions of the instances of other classes which are visually close to the ground truth class\footnote{In Figure \ref{fig:true}, the second largest prediction made by PtR is at another class of similar appearance; in Figure \ref{fig:false}, the wrongly predicted class is also similar to the ground truth.}. 
This indicates that the pseudo task regularizer implicitly helps the network focus on and learn to distinguish classes of higher visual similarity, 
and hence helps the classification accuracy overall (for around 1\% on CUB200's validation set). 
At the same time, from our observations, PtR does not tend to produce so many minor probabilities on other classes which are less similar as the baseline model does.
This aids the regularized model to suppress the uncertainties and focus on a few most similar candidate classes.

\section{Related Work}
\label{sec:relate}

Regularization methods that are commonly used for fine-tuning CNNs can be generally grouped into four categories: data perturbation, parameter norm penalty, dropout, and multi-task learning. 
Image augmentation as a popular form of image perturbation has been proven to be  particularly useful to prevent CNNs from overfitting. 
In this paper, 
we also assume to perturb our training instances 
by random augmentations. 
The supervision signal can be perturbed for better regularization as well. 
This can be achieved by learning to predict soft targets rather than hard binary ones as in \cite{disturblabel}. 
In this work, label perturbation is not considered so that we can deliver more ablated studies on the effectiveness of auxiliary training objectives. 

The parameter norm penalty, or {\it weight decay} more specifically, has been one of the most common ways of regularization in training deep models. 
Our PtR leveraged weight decay by default, but we also evaluated PtR without weight decay to study its impact on accuracy.
Another method being apparently similar to weight decay is 
to use feature norm penalty (FPN) or feature contraction
on the representation layer of a network 
\cite{fewshot,LiM2018,fc2019}.
Superficially, FNP would resemble to PtR if the regression target for PtR was towards a static norm of zero without involving randomness. (An additional feature of PtR which should be noted is that it also balances objectives automatically. See Section \ref{sec:vargrad} for the technical difference.)
{\it Dropout} is also one of the standard techniques to improve model regularization by temporarily shielding a part of the hidden units in the bottleneck layer and fully connected layers during training \cite{dropout,cdrop}. 
For the VGG-16 structures which we employed in our experiments, dropout was also used after flattened hidden layers. 

A more recent approach of regularization in transfer learning has been to train CNNs with an auxiliary task/objective through multi-task learning \cite{LwF,Borrow}.
These objectives in the recent best performing methods are often designed with the expectation that more generic features are less likely to overfit to the target task in a few different ways.
In \cite{Borrow}, the network was simultaneously trained by the target data and a number of selected source data samples that are similar to the target data 
when viewed in low-level features.  
Another way to encourage CNN regularization, instead of relying on the availability of foreign data, was to let the model stay not too far from the original structure of the model trained by a large source task. 
As demonstrated by \cite{LwF}, one can attempt to retain the predictions of the target domain images made by the off-the-shelf source model while learning on the new target task
(we acknowledge that the original intention of doing so in \cite{LwF} was not for regularization).
It is therefore reasonable to interpret the use of off-the-shelf predictions as implicitly using the source domain training data. 
The other way to make the trained models attracted to the original model was to explicitly force the target model being trained to stay in a vicinity of the source model in the weight space; the work in \cite{IndB} leveraged the inductive bias in a fine-tuning scenario to prevent the learned features from becoming overly specific. 

For fine-grained vision tasks, a very recent approach \cite{PC} suggested to ``confuse'' the network by encouraging different class-conditional probability distributions to come closer together, thus reducing the inter-class distance. 
In the cases where only style transfer is considered, one can attempt to reduce the domain variance through certain metrics (i.e., perform domain adaptation) as in \cite{Tzeng_2015_ICCV,pmlr-v37-long15}.

A common aspect of these aforementioned regularizers is that they depend on a concrete task or objective.
But since they are not purposed for explicitly optimizing the target objective, they can all be largely seen as ``distractors''.
The approach studied in this paper also causes distractions, but we suggest that a distractor could work equally effectively without involving a concrete auxiliary objective or any form of source domain data as supervision labels.

\section{Conclusions and Discussions}

We have introduced a {\it Pseudo-task Regularization} (PtR) which leverages a multi-task learning framework to generate useful interference by a pseudo-regression task, for improving regularization for transfer learning with limited data samples. 
The regularization effect from PtR is dynamic in that PtR adjusts the strength of the regularization based on the gradient norms of the target objective and the pseudo-task.
Unlike existing approaches, PtR does not depend on a concrete or real regularization objective. 
Surprisingly, we observed that the performance gain brought about by the simple PtR was on a par or better than the related recent solutions, 
and therefore we suggest that PtR 
is available
as an efficient alternative to recent best performing regularization methods that are based on concrete objectives.


We attribute the generalization gain from PtR to two aspects while a widely recognized theory on DNN generalization is yet to be established \cite{3factors, anisotropic}. 
First, the gradient noise generated by PtR in SGD makes batch-wise gradients noisier and results in smaller equivalent batch size in that respect. 
The resulting increased quotient of learning rate and (equivalent) batch size have been shown helpful to escape sharp minima \cite{3factors,minima}.
The nosier gradients prolongs training which could in turn encourage networks exploring more and better local minima, hence class disambiguation. 
Second, the superiority of anisotropic noise of SGD over its isotropic counterpart has been demonstrated \cite{anisotropic}, and the anisotropic gradient noise generated by PtR could increase the chance to find flatter minima.
 

Furthermore, it will be also interesting to study the impact of PtR on the calibration performance of the network \cite{ece}
, and how PtR's gradients interact with those of the baseline as in \cite{UCR}.
This is however beyond the scope of the current work as well as the above mentioned relation to the loss landscape -- those are among the topics for future research.

\section*{Acknowledgement}
\vspace{-1mm}
\noindent
We would like to thank Ryuzo Okada and colleagues of Toshiba Corporate R\&D Center for funding the research and for collaboration. We also wish to thank Vladimir Li and Matteo Gamba for fruitful discussions, and NVIDIA Corporation for their generous donation of GPUs. The second author is supported by the Swedish Research Council, which is gratefully acknowledged.


{\small
\bibliographystyle{ieee_fullname}
\bibliography{egbib}
}

\end{document}


\title{Regularizing CNN Transfer Learning with Randomised Regression}

\author{Yang Zhong \qquad Atsuto Maki\\
Division of Robotics, Perception and Learning \\
KTH Royal Institute of Technology\\
{\tt\small \{yzhong, atsuto\}@kth.se}
}

\maketitle


%
\section{The scales of the datasets used in the paper}
\begin{figure}[H]
\centering 
\includegraphics[width=0.4\textwidth]{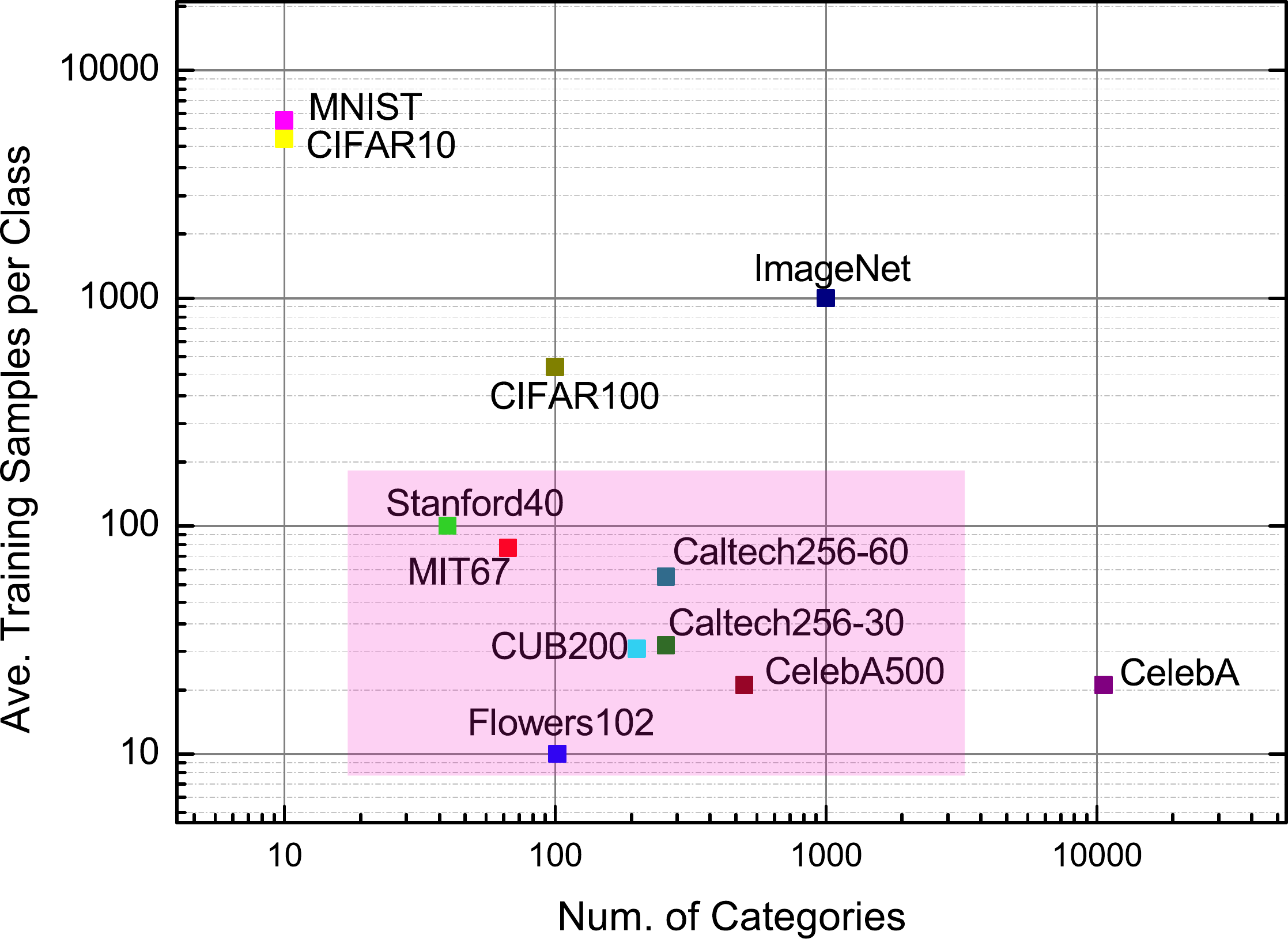}
\caption{\it The scales of datasets used in our experiments (in the magenta shadowed area) among other popular transfer learning benchmarks, e.g. CIFAR10 and MNIST, as well as the commonly used source datasets such as ImageNet \cite{imagenet_cvpr09} and CelebA datasets \cite{CelebA}. 
The average number of per-class training instances (vertical-axis) and the total number of classes (horizontal-axis) of each dataset are shown in $log_{10}$ scale. 
}
\label{fig:db}
\end{figure}

Our paper demonstrates and discusses a ConvNet regularizer which works without extra knowledge or data (unlike \cite{Borrow,LwF,IndB,PC}) on several transfer learning tasks for image classification.
These tasks cover different scenarios of transfer learning including fine-grained object classification and close-set human face identification.
They contain more sparse training samples than some commonly studied transfer learning tasks, e.g. CIFAR \cite{cifar} and MNIST \cite{mnist}.
The total number of classes and the number of training sampler per class of these datasets are compared in Figure \ref{fig:db}.
It can be seen that the transfer learning tasks we study in this paper feature either a few samples per class or many categories with limited training samples; the total training samples are fewer than $6000$ for the selected datasets except ``CelebA-500'' while its average training samples per class is lower than most of the selected datasets.

\label{sec:introdu}

\section{Timing to bring in PtR}
The pseudo regression task is brought into the training pipeline at a proper stage, and the timing was settled by considering the training efficiency and model effectiveness. 
That is, our pseudo-task branch is enabled to jointly train the network when the entire network is considered as relatively close to convergence
\footnote{This is different from multiple training stages as in \cite{LwF} where the additional imposed regularizer starts to train the network only after the target classifier is converged on the target task.}.

\begin{figure}[h]
\centering 
\includegraphics[width=0.4\textwidth]{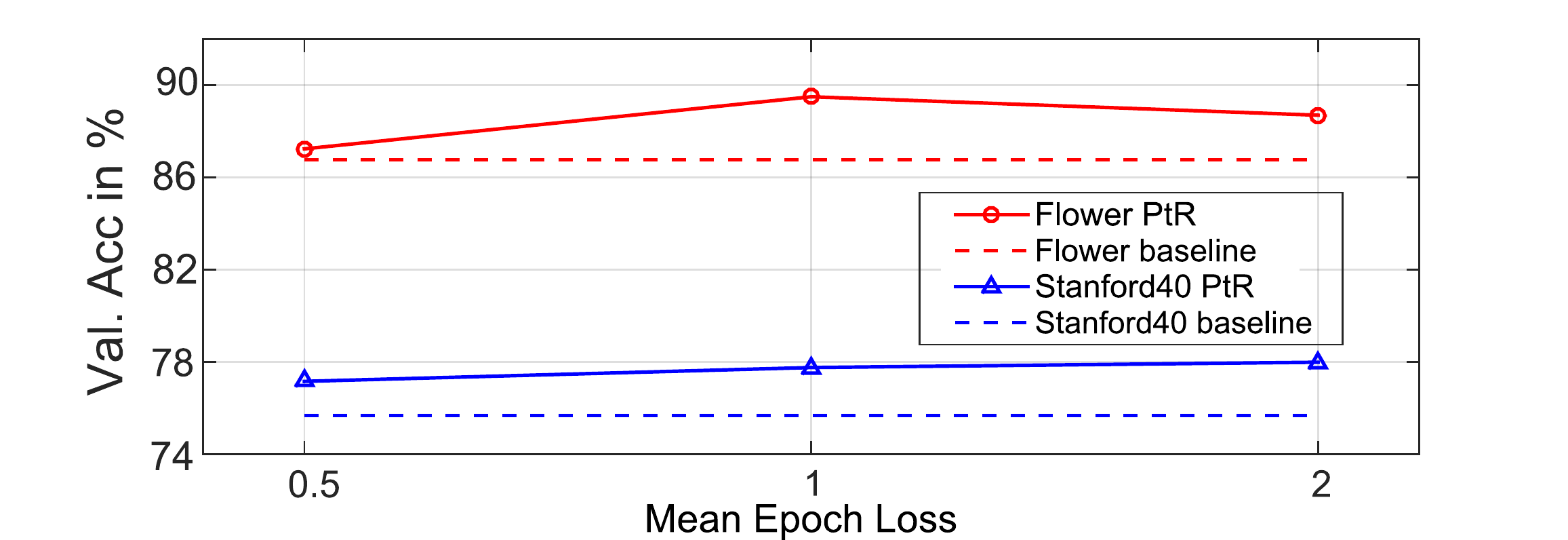}
\vspace{-.8em}
\caption{{\it Validation accuracy (in \%) on Flower and Stanford40, according to different timing when PtR joins the network training on the course of model convergence using a varying threshold, $T$, for the mean epoch cross-entropy loss (with SML1 regression).}}
\label{fig:tvalues}
\end{figure}

As shown in Figure \ref{fig:tvalues}, we experimented with a varying mean epoch training loss ($T$) on Flower102 and Stanford40 to determine a practical one for efficient training and models' effectiveness.
It can be observed that 
PtR improves on the baseline accuracies (fine-tuning) in all cases and 
the PtR at $T$=1 resulted in a higher accuracy on average. 
We chose $T$=1 as a practical setting in all the rest of experiments since a larger $T$ does not contribute to the model effectiveness while negatively prolonging the training time.


\section{Standard Deviations of PtR with ResNet}
The average classification accuracies (in \%) with standard deviations of the pseudo-task regularization (PtR) with the ResNet architectures are given in Table \ref{tab:std} to provide more details for Section 3 in the main manuscript to compare to \cite{PC,IndB}. 

\begin{table*}[ht]
\centering
\small
\caption{The ResNet test accuracies, with standard deviations given in parenthesis, on the datasets used in our work. }
\begin{tabular}{ccccc}
\toprule
              & \multicolumn{2}{c}{ResNet-50} & \multicolumn{2}{c}{ResNet-101} \\ 
              & Baseline       & PtR           & Baseline       & PtR            \\ \midrule
CUB200        & 80.3\% (0.19)    & 81.9\% (0.23)   & -              & -              \\ 
Flower102     & 91.0\% (0.32)    & 91.8\% (0.22)   & 90.6\% (0.17)    & 91.6\% (0.26)    \\ 
MIT67         & 77.4\% (0.72)    & 77.9\% (0.65)   & 78.7\% (1.10)    & 77.9\% (0.63)    \\ 
Caltech256-30 & -              & -             & 84.0\% (0.15)    & 84.5\% (0.19)    \\ 
Caltech256-60 & -              & -             & 86.8\% (0.18)    & 87.2\% (0.07)    \\ \bottomrule
\end{tabular}
\label{tab:std}
\end{table*}


\section{Impact of weight decay}
Weight decay may have some observable impact on the performance of the fine-tuning baseline, but it can hardly impact the performance of PtR. 
We collect all the related results and show in Table \ref{tab:wd} \footnote{All the numerical results in the supplementary document were averaged from five independent runs.}.

\begin{table}[h]
\centering
\small
\caption{Impact of weight decay in fine-tuning (FT) and PtR (ResNet50).
``w/'' and ``wo/'' stands for with default weight decay settings and without using weight decay respectively.
}

\begin{tabular}{ccccc}
\toprule
              & FT w/    & FT \textbf{wo/}    & PtR w/    & PtR \textbf{wo/}    \\ \midrule
CUB200        & 80.3\%   & 81.0\%    & 81.9\%    & 82.0\%     \\ \hline
Caltech256-30 & 83.2\%   & 83.2\%    & 83.9\%    & 83.9\%     \\ \hline
Caltech256-60 & 86.6\%   & 86.7\%    & 87.1\%    & 87.0\%     \\ \bottomrule
\end{tabular}
\label{tab:wd}
\end{table}

\section{Training from scratch}
We have also evaluated the PtR when training models from scratch as shown in Table \ref{tab:frmsch}. 
Comparing the regularization gain in Table 2 and Table 5 in the main paper, it can be observed that PtR demonstrates stronger regularization compared to the scenarios where the ImageNet initialization was used. 
It can also be noticed that the gain on the Caltech256 is consistently smaller than that of the CUB200 in both scenarios. 
Considering the slightly larger category on the Caltech256 (see Figure \ref{fig:db}), it suggests that training with more real samples leaves lesser room for PtR to improve regularization.

\begin{table}[h]
\centering
\small
\caption{Comparing test accuracy of fine-tuning (FT) baseline and PtR with ResNet50 on the CUB200 and Caltech256-30 when training from scratch. The absolute performace gain in \% is listed in the last column for ease of comparison.}
\begin{tabular}{cccc}
\toprule
              & FT      & PtR     & Gain   \\ \midrule
CUB200        & 57.33\% & 61.71\% & 4.38\% \\ \hline
Caltech256-30 & 46.21\% & 48.25\% & 2.04\%  \\ \bottomrule
\end{tabular}
\label{tab:frmsch}
\end{table}

\section{Disentangle batch normalization}
Batch normalization has been commonly used as a regularization module in network training. 
To explore the interaction of its regularization effect with PtR, we applied it to off-the-shelf VGG-16 models with batch normalization layers.
It is a good candidate for this study is because it features a more uniform convolutional-pooling structure that facilitates disentangling other factors (compared to ResNet for example) and is sufficiently deep.
The comparative results are shown in Table \ref{tab:vgg16bn}.

Comparing to Table 2 in the main paper, it can be found that batch normalization improved the baseline for around 1.6\%.
The use of batch normalization seems to improve the PtR: 
together with batch normalization, PtR achieved an extra 1\% regularization gain compared to that in Table 2. 

\begin{table}[h]
\centering
\small
\caption{Comparing PtR to fine-tuning (FT) on the CUB200 using the ImageNet pre-trained VGG-16 with batch normalization layers.}
\begin{tabular}{cccc}
\toprule
       & FT      & PtR     & Gain   \\ \midrule
CUB200 & 76.70\% & 80.73\% & 4.03\% \\ \bottomrule
\end{tabular}
\label{tab:vgg16bn}
\end{table}

 
\section{Case studies on the effect of the Psudo-task Regularization (PtR)}
We provide ten examples (cf. Section 5 of the paper) with details as to how PtR impacts the probability predictions in comparison to the corresponding responses from the baseline networks trained by the standard finetuning.
Specifically, we compare how the predictions of PtR differ from the baseline in four scenarios:
\begin{enumerate}
\item PtR correctly rectifies finetuning's errors (i.e., true rectification);
\item PtR gives wrong predictions while the predictions by finetuning were right (i.e., false rectification);
\item Both PtR and finetuning give correct predictions (i.e., both correct);
\item Neither PtR nor finetuning gives correct predictions (i.e., both wrong).
\end{enumerate}

\begin{figure*}[]
    \centering
    \begin{subfigure}[]{0.95\textwidth}
        \includegraphics[width=\textwidth]{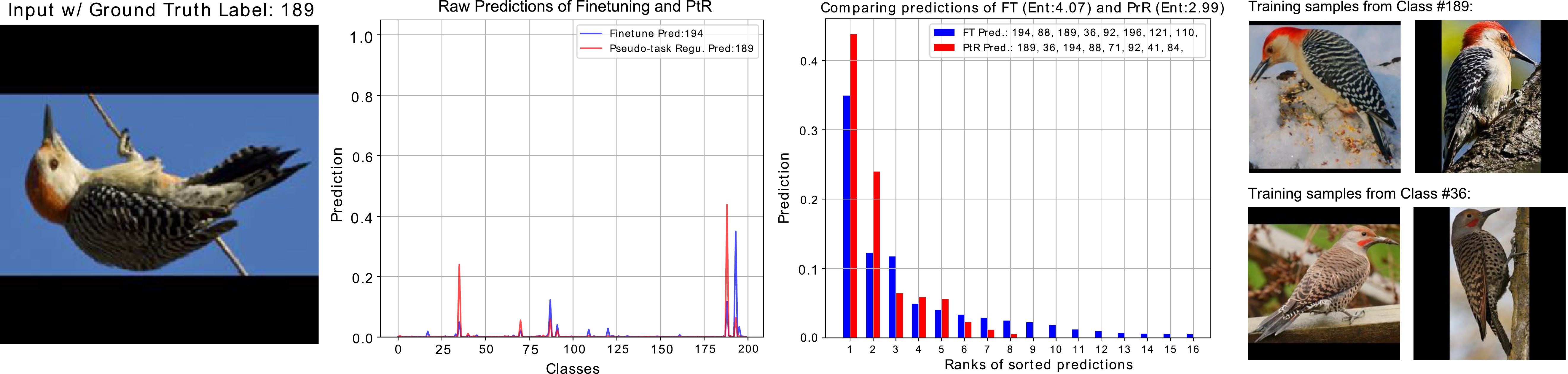}
        \caption{\it PtR gives correct rank-1 class prediction and it is more certain about it than FT is about the misclassified result for the displayed sample. 
        		In addition, PtR also gives a relatively high score on the second prediction which is a class quite similar to the ground truth (compare patterns on the wings and colors of head of the training examples from Class 36 to Class 189). 
                Only eight predicted major probabilities (those higher than 0.5\%)  are made while finetuning generates twice as many; the entropy of PtR turned out to be less than that of FT.}
        \vspace{1.5em}
        \label{fig:1a}
    \end{subfigure}
    ~ 
    \begin{subfigure}[]{.95\textwidth}
        \includegraphics[width=\textwidth]{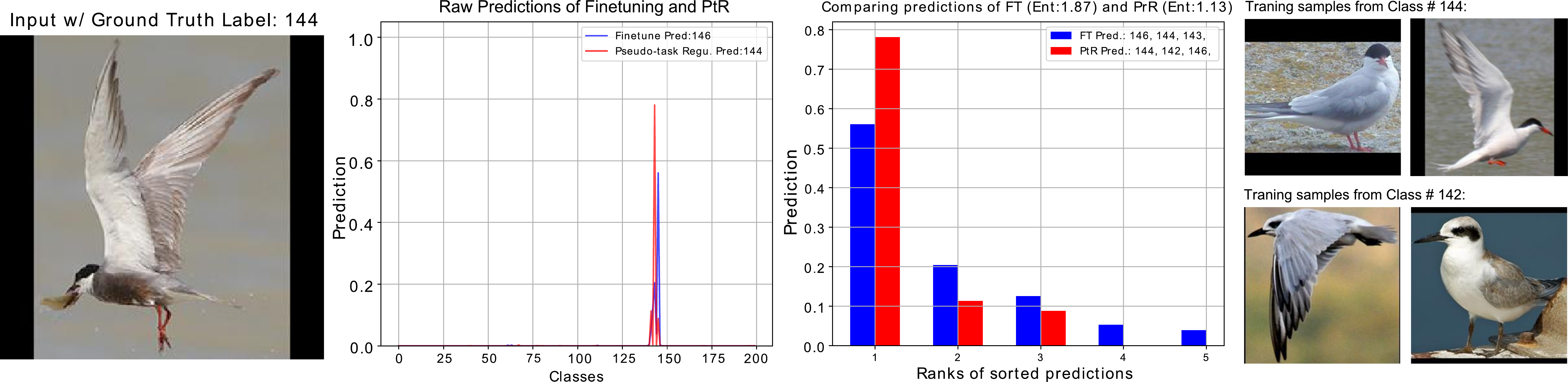}
        \caption{\it PtR is quite confident with rank-1 prediction and gives lower but still significant scores to visually similar classes as a result. Similarly to the example in (a), major prediction scores are given to only three classes by PtR whereas FT picks two more classes. The entropy of PtR is also lower than that of FT.}
        \vspace{1.5em}
        \label{fig:1b}
    \end{subfigure}
    ~ 
    \begin{subfigure}[]{.95\textwidth}
        \includegraphics[width=\textwidth]{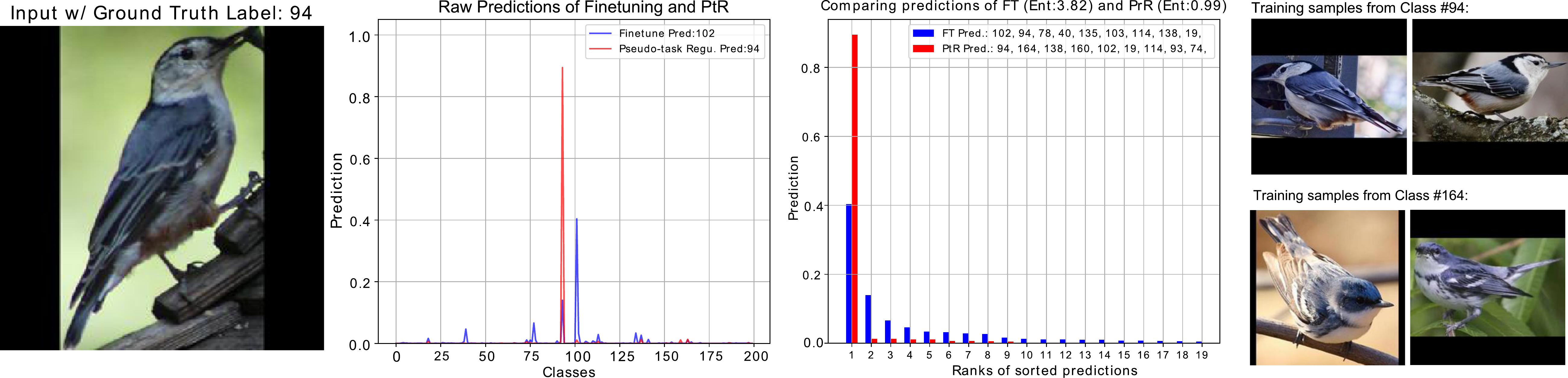}
        \caption{\it PtR gives the highest prediction score to the input image (among all of the three examples in this True Rectification scenario) and the prediction scores on other classes are quite marginal. But it can also be seen that PtR still tends to identify fewer similar classes given that only eight other classes receive a score around 1\% while finetuning selects ten more classes.}
        \vspace{1em}
        \label{fig:1c}
    \end{subfigure}
    \vspace{-.5em}
    \caption{\it Examples of True Rectifications.}
    \label{fig:1}
\end{figure*}

\begin{figure*}[]
    \centering
    \begin{subfigure}[]{0.95\textwidth}
        \includegraphics[width=\textwidth]{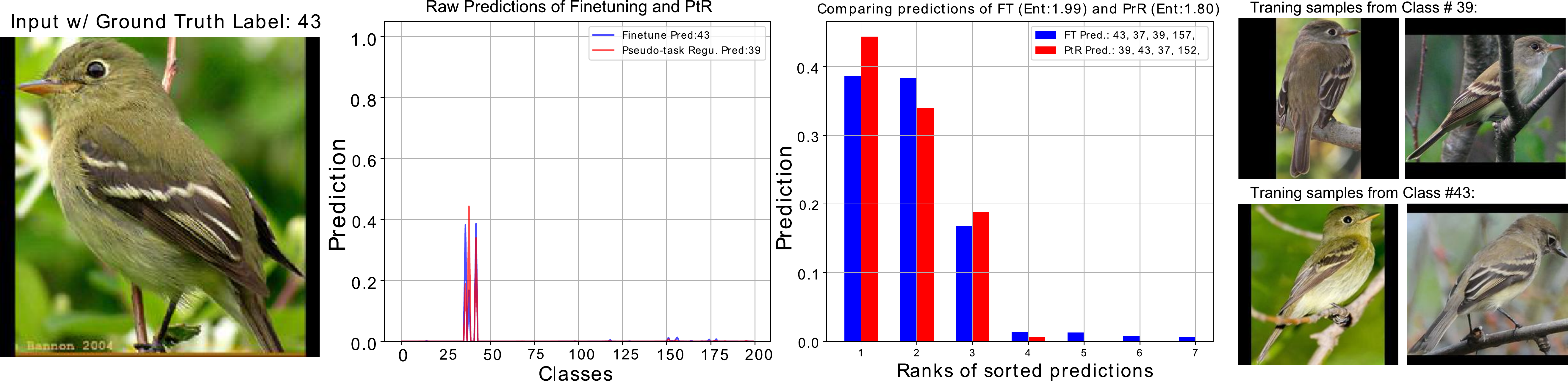}
        \caption{\it This is a hard positive example to finetuning on which PtR makes an error. From the predictions (see the 3rd grid), it can be seen that both methods identify the similar classes --- Class 37, 39, and 43; FT generates a slightly higher score on the correct class. PtR again focuses on only four classes and the entropy of the prediction scores is slightly lower.}
        \vspace{2em}
        \label{fig2:a}
    \end{subfigure}
    ~ 
    \begin{subfigure}[]{.95\textwidth}
        \includegraphics[width=\textwidth]{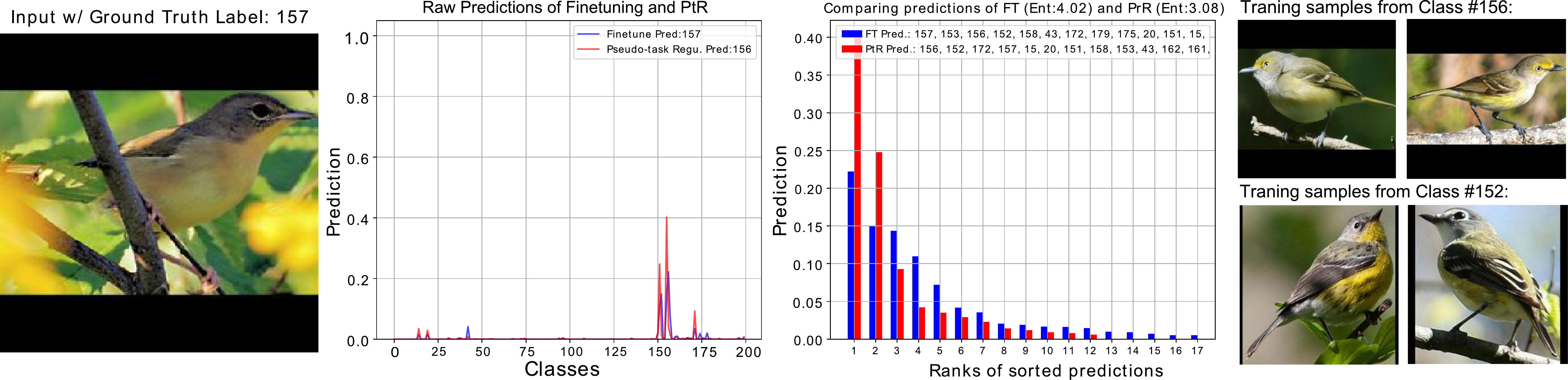}
        \caption{\it PtR gives more confidence than FT does for the top-2 predictions and it consequently assigns lower scores on other similar categories. The number of major predictions are fewer than that of FT. In this way, it can focus on a few similar classes and generates predictions with a lower entropy. }
        \vspace{2em}
        \label{fig2:b}
    \end{subfigure}
    ~ 
    \begin{subfigure}[]{.95\textwidth}
        \includegraphics[width=\textwidth]{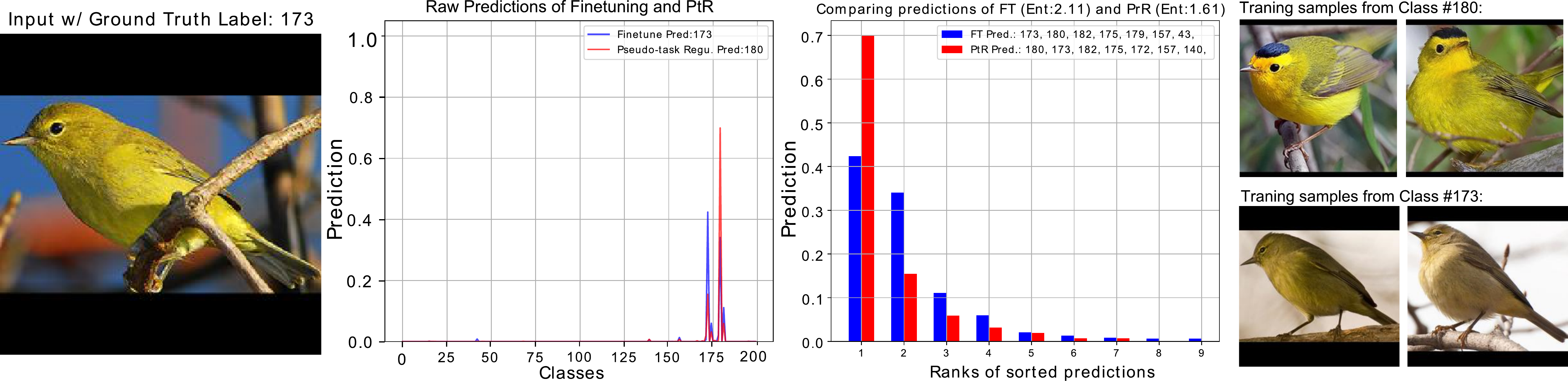}
        \caption{\it Although PtR gives high confidence on a similar but wrong class, it still identifies the correct class in the rank-2 prediction. The number of major predictions are slightly fewer than that of FT and they are less noisy as indicated by the lower entropy.}
        \vspace{1em}
        \label{fig2:c}
    \end{subfigure}
        \caption{\it Examples of False Rectifications.}
        \label{fig:2}
\end{figure*}

\begin{figure*}[]
\vspace{-9.5em}
    \centering
    \begin{subfigure}[]{0.95\textwidth}
        \includegraphics[width=\textwidth]{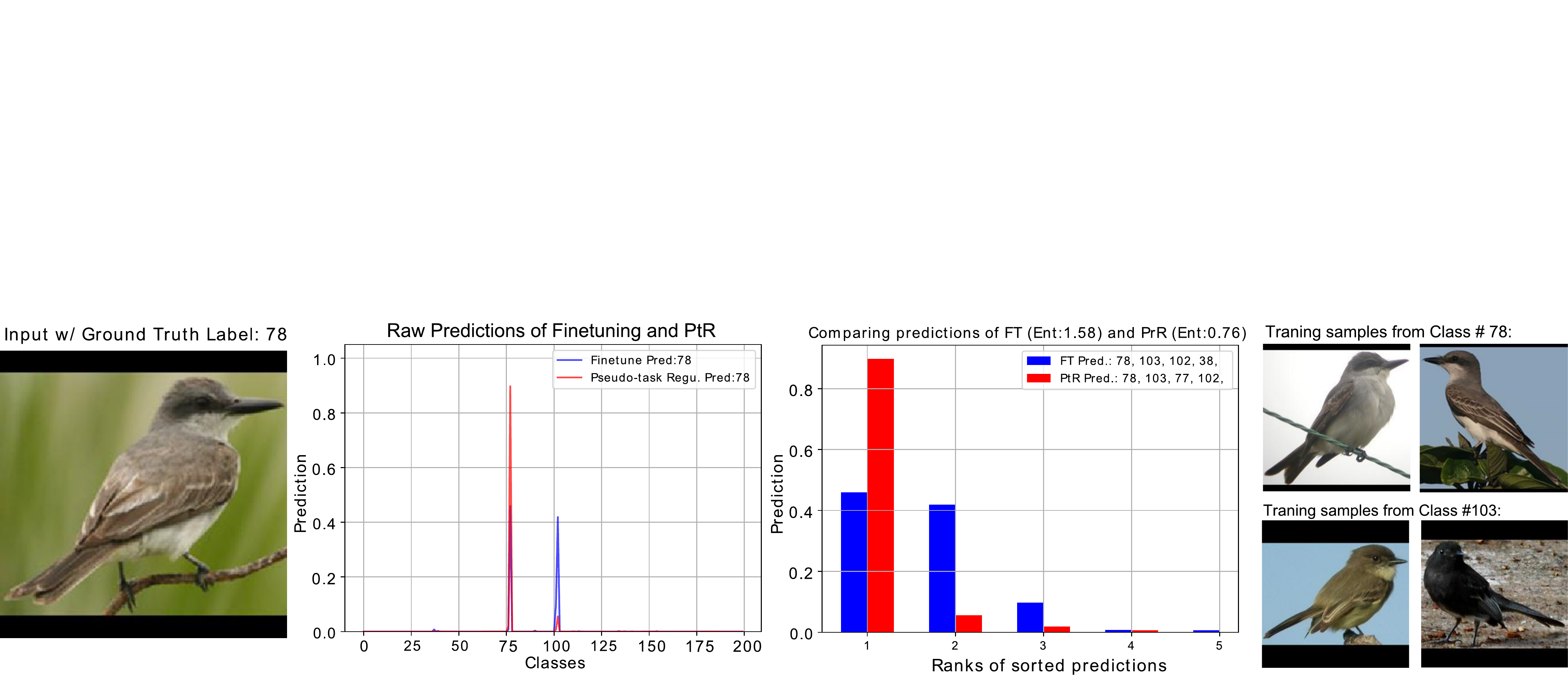}
        \caption{\it Another hard positive example for FT. It can be seen that the predictions made by PtR are more discriminative than those by FT, but at the same time PtR can still correctly identify the visually correlated Class 103. The number of major predictions of PtR is fewer than that of FT; the entropy of the PtR's prediction is also lower than FT.  }
        \vspace{1.5em}
        \label{fig3:a}
    \end{subfigure}
    ~ 
    \begin{subfigure}[]{.95\textwidth}
        \includegraphics[width=\textwidth]{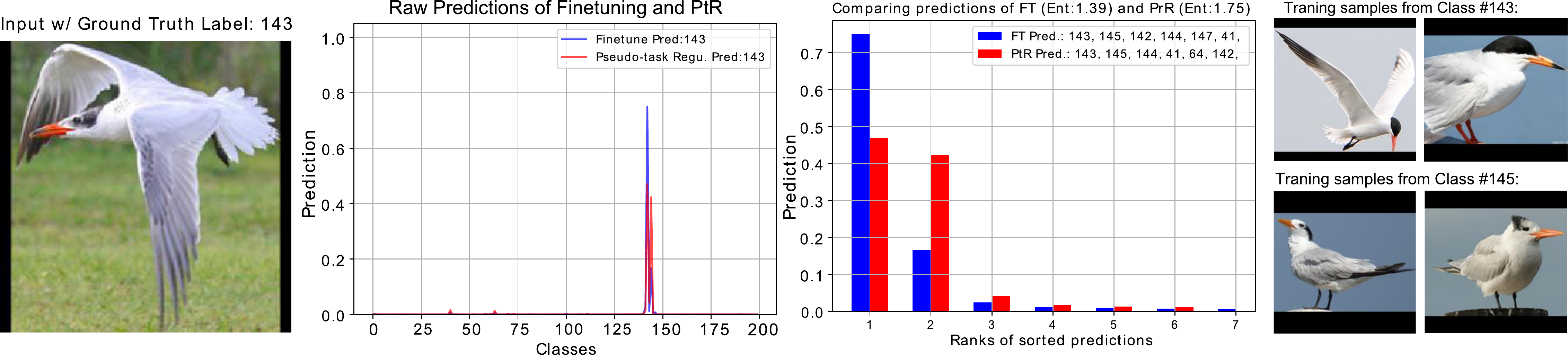}
        \caption{\it Although PtR still correctly identifies the object, the close scores (of rank-1 and rank-2 predictions) suggest that it also suspects the rank-2 predicted class. This means that the model trained by PtR has focused mostly on the first two predicted classes. The predictions of PtR are slightly noisier than FT, but the number of major probabilities is one fewer than that of FT.}
        \vspace{.5em}
        \label{fig3:b}
    \end{subfigure}
   \vspace{-.6em}
   \caption{\it Examples of Correct Predictions by Both Methods.}
   \label{fig:3}
   \vspace{1.5em}
%
%
%
    \centering
    \begin{subfigure}[]{0.95\textwidth}
        \includegraphics[width=\textwidth]{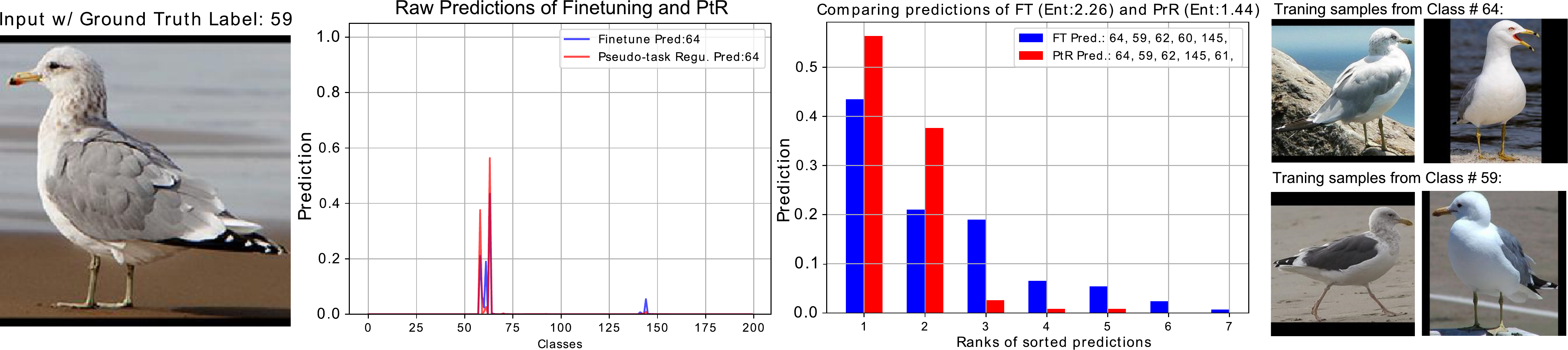}
        \caption{\it Both FT and PtR make correct predictions at the second rank. PtR still predicts fewer minor probability and makes less noisy predictions than FT.}
        \vspace{1.5em}
        \label{fig4:a}
    \end{subfigure}
    ~ 
    \begin{subfigure}[]{.95\textwidth}
        \includegraphics[width=\textwidth]{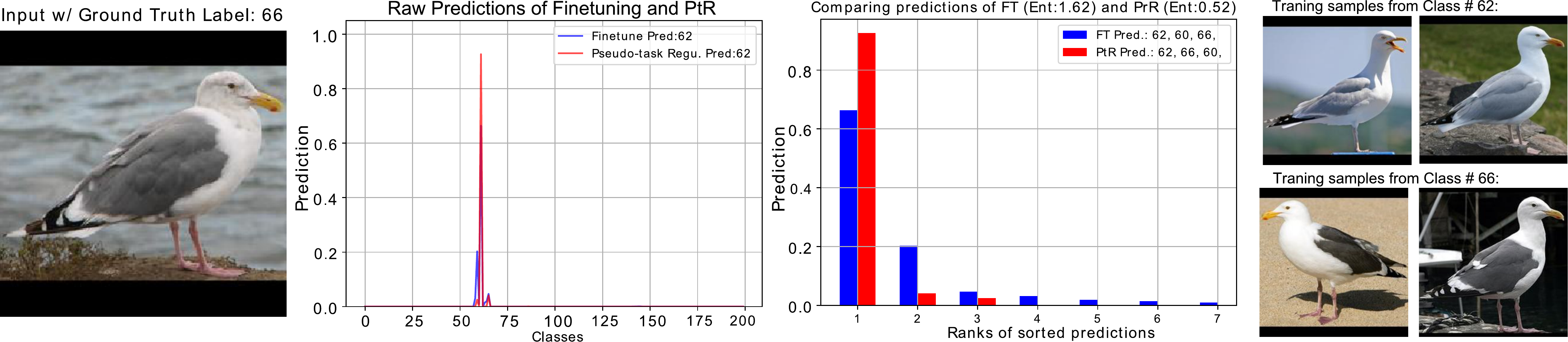}
        \caption{\it Although both rank-1 predictions are wrong, PtR correctly identifies the right class at the rank-2 prediction. As PtR gives confidence in its rank-1 prediction, the rest of the identified classes receive lower scores. PtR in this case still produces fewer minor probabilities than FT.}
        \vspace{.5em}
        \label{fig4:b}
    \end{subfigure}
    \vspace{-.6em}
    \caption{\it Examples of Wrong Predictions by Both Methods.}
    \label{fig:4}
\end{figure*}

Figure \ref{fig:1}, \ref{fig:2}, \ref{fig:3}, and \ref{fig:4} show the case studies corresponding to each scenario, respectively. 
In each of them, input images following the scenario are shown at the left most, with the ground truth class labels on the top. 
Next to those, the raw probabilities given by PtR and finetuning are displayed in the grid. 
The predictions by the two methods are then sorted according to the magnitude, respectively, and we show the sorted probabilities which are larger than a small value of 0.5\%, which we call major probabilities, in the third grid.
The predicted labels (both by finetuning and PtR) are given in the legend, and the entropy of each distribution is provided on top.
``Finetuning'' and ``predictions'' are denoted as ``FT'' and ``Pred'', respectively.

In the right most, four images from the training set are shown for reference:
two of them come from the rank-1 predicted class of PtR and another two images come from the second predicted class.
This allows us, by comparing those samples and the sorted probabilities, to subjectively observe how the visual similarities have been valued in the predictions by PtR more easily.

\vspace{1.5em}

\vspace{1.5em}

Throughout the case studies demonstrated in the figures, an important characteristic of PtR can be found: with PtR the network tends to identify fewer classes as being similar to the input class. 
In particular, many of the minor probabilities (predicted as low values) are eliminated compared to the original distributions from the finetuning baseline.
As a result, the predictions of PtR are in general less noisy than those by finetuning.  
We suspect that noisy predictions by finetuning are caused by overfitting; our case studies 
suggest that PtR is capable of alleviating such a problem in a transfer learning scenario.
Besides, PtR often encourages the predictions of visually similar classes in various scenarios.

As well as the qualitative case studies provided so far, we also summarize some statistics of our case studies in addition to Section 5 in the main manuscript. 
The validation set of CUB200 contains 584 randomly selected images. 
The classification accuracy for finetuning and PtR are 79.8\% and 80.8\%, respectively. 
The number of true rectification is 27 and the number of false rectification is 21. 
PtR and finetuning make correct predictions on a common set of 445 images and they both make wrong predictions on 91 images.

\balance

{\small
\balance
\bibliographystyle{ieee}
\bibliography{egbib}
}